\pdfoutput=1

\documentclass[11pt]{article}

\usepackage{acl}

\usepackage{times}
\usepackage{latexsym}

\usepackage[T1]{fontenc}

\usepackage[utf8]{inputenc}

\usepackage{microtype}
\usepackage{float}
\usepackage{booktabs}
\usepackage{multirow}
\usepackage{amsmath}
\usepackage{amssymb}
\usepackage{subcaption}
\usepackage{algorithm}
\usepackage{algorithmic}
\usepackage{textcomp}
\usepackage{graphicx}
\usepackage{colortbl}

\DeclareMathOperator*{\argmax}{arg\,max}

\definecolor{mygray}{gray}{.9}

\title{$DG^2$: Data Augmentation Through \\ Document Grounded Dialogue Generation}

\author{
    Qingyang Wu\textsuperscript{\rm 1} 
    Song Feng\textsuperscript{\rm 3} 
    Derek Chen\textsuperscript{\rm 2} 
    Sachindra Joshi\textsuperscript{\rm 3}  
    \textbf{Luis A. Lastras\textsuperscript{\rm 3} 
    Zhou Yu\textsuperscript{\rm 1}} \\
    \textsuperscript{\rm 1}Columbia University \,
    \textsuperscript{\rm 2}ASAPP \,
    \textsuperscript{\rm 3}IBM Research AI \\
    \{qw2345, zy2461\}@columbia.edu, dchen@asapp.com \\
    \{ sfeng@us, jsachind@in, lastrasl@us\}.ibm.com \\
}

\begin{document}
\maketitle
\begin{abstract}

%

Collecting data for training dialog systems can be extremely expensive due to the involvement of human participants and need for extensive annotation.
Especially in document-grounded dialog systems, human experts need to carefully read the unstructured documents to answer the users' questions.
As a result, existing document-grounded dialog datasets are relatively small-scale and obstruct the effective training of dialogue systems.
In this paper, we propose an automatic data augmentation technique grounded on documents through a generative dialogue model. The dialogue model consists of a user bot and agent bot that can synthesize diverse dialogues given an input document, which are then used to train a downstream model.
When supplementing the original dataset, our method achieves significant improvement over traditional  data augmentation methods.  We also achieve great performance in the low-resource\vphantom{and unseen document} setting.

\end{abstract}

\section{Introduction}

Most of human knowledge is stored in the form of documents.
Those documents not only help people to find answers to factoid questions like when George Washington was born, but also provide instructions for tasks such as how to assemble a desk bought from IKEA.
How to comprehend and retrieve information from documents is a challenging research problem for dialog systems.
As this task has real-world applications, there have been many works \cite{DBLP:conf/emnlp/RajpurkarZLL16,DBLP:conf/acl/RajpurkarJL18,2019-tom-nq,DBLP:conf/emnlp/YangYM15} trying to tackle this challenge.

Recently, contextual and dialog-based question answering systems have gained more research attention and are often referred to as document-grounded dialog systems \cite{DBLP:journals/corr/abs-2004-13818}.
Early works such as QuAC \cite{choi18quac} and CoQA \cite{reddy19coqa} first explored the direction of contextual question answering.
Later, MANtIS \cite{DBLP:journals/corr/abs-1912-04639} and DoQA \cite{campos20doqa} incorporated the consideration of user intents to have more natural and coherent conversations, and ShARC \cite{saeidi18sharc} added follow-up questions for the agent.
Compared to previous datasets, Doc2Dial \cite{feng20doc2dial} further expands on the number of scenes and domains, which require the model to have a stronger capability in seeking 
information within the document. 

\begin{figure}[t]
    \centering
    \includegraphics[scale=0.4]{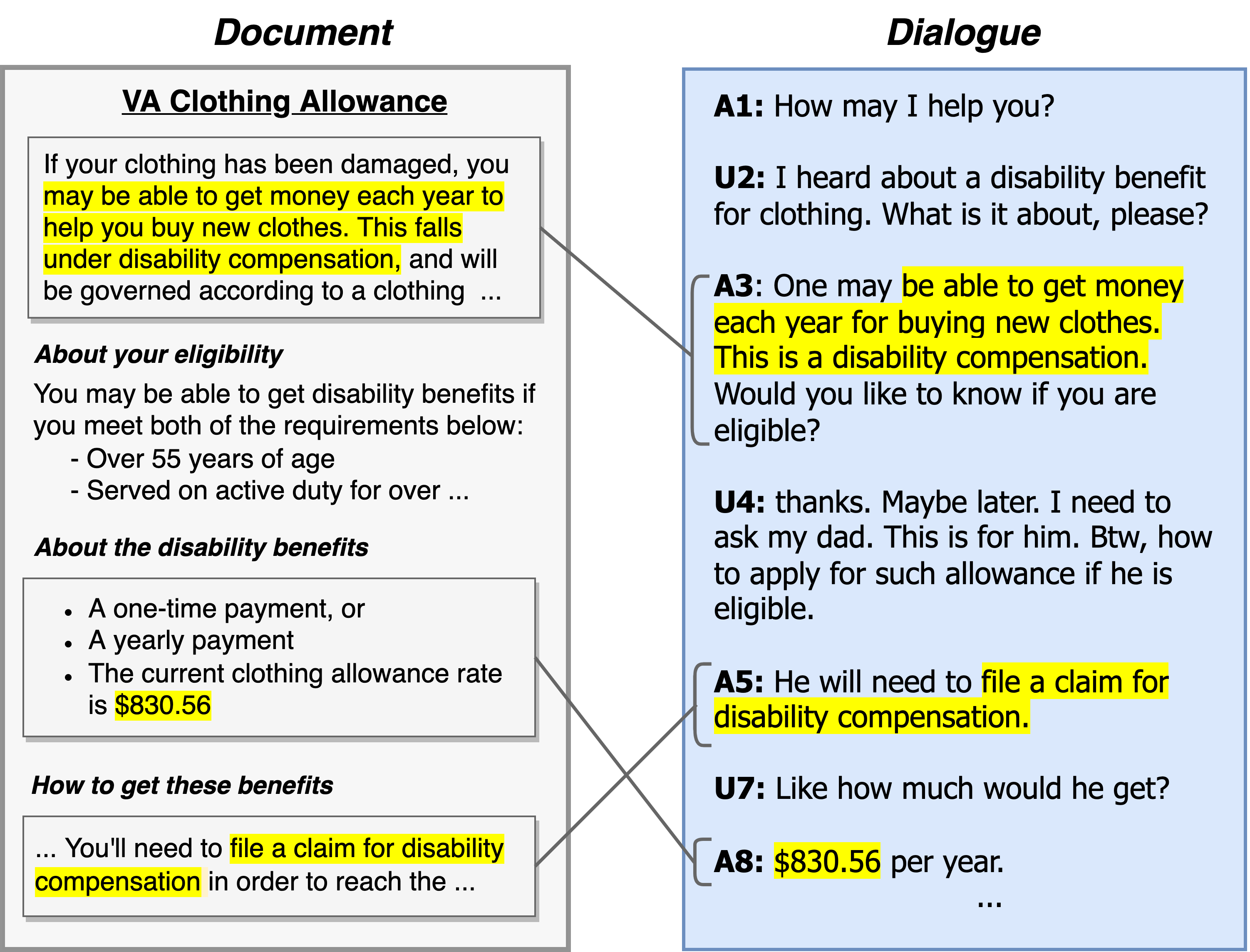}
    \caption{An example from Doc2Dial of dialogue conversation produced from grounding to an associated document. The agent must select the correct spans and engage in a fluent manner to generate a proper response.} 
    \label{fig:intro}
\end{figure}

However, as the relations between conversations and documents become more complex, 
the cost of collecting large-scale datasets also becomes more expensive.
As a consequence, one main obstacle for developing effective document grounded dialog systems is the lack of sufficient data.
In chit-chat scenarios, recent works such as DialoGPT \cite{DBLP:conf/acl/ZhangSGCBGGLD20}, Meena \cite{adiwardana20meena}, and Blender \cite{roller21blender} have achieved human-like performance by taking the advantage of training on a large-scale corpus.
Similarly, task-oriented dialog systems such as ARDM \cite{wu21ardm} 
and SimpleTOD \cite{hosseini20simpletod} have also utilized large-scale corpora or pre-trained models to achieve good performance. 
The aforementioned models were trained with millions of samples, while the current document-grounded dialogue datasets like Doc2Dial \cite{feng20doc2dial} only contain thousands of conversations.
Training on such a small-scale dataset constrains the performance of neural network models.  
Therefore, augmenting existing datasets can help build more effective document-grounded dialogue system.

One popular approach to augmenting datasets is to paraphrase existing seed data.
The most straightforward form of paraphrasing is to directly use a model trained to generate paraphrase pairs \citep{gao20paraphrase}. Back-translation serves as another type of paraphrasing, which first translates a sentence into another language and then  back again \citep{chadha19bertqa,bornea21translation}. Back-translation ensures quality and correctness of the augmented data and often shows improvement in downstream models.  Both methods aim to provide variety to the training data without altering the semantics of the original sentences.
However, these methods only operate on the existing dialogue data and fail to take advantage of the available document for augmentation.


Another direction for data augmentation is to generate examples from scratch by grounding to auxiliary documentation. \citet{lewis21paq} generate question-answer pairs with a model pre-trained on available training data.  This often requires additional filtering or denoising measures to ensure correctness of generated data. Also, these models are built for the purposes of single-turn question answering, rather than multi-turn dialogues. 

Inspired by \citet{alberti19synthetic}, we propose an automatic document-grounded dialogue generation ($DG^2$) method that augments the amount of data available for training a dialogue system. The model consists of a user bot and a agent bot that alternately generates utterances to complete a conversation.
The user bot includes a span extraction model that can first select a passage and then predict the rationale start and end positions inside a passage.
The agent bot has a denoising mechanism to filter out generated rationales irrelevant to the conversation.  
The user bot behaves as a teacher, and begins by selecting a passage from the document that is most relevant to the current context.  It then selects a rationale span from this passage and generates the user utterance.
The agent bot behaves as a student.  It first checks if it can find the correct rationale span, and then generates the agent response.  This process repeats until an entire dialogue is generated.

We evaluate our model on a representative document-grounded dialog dataset Doc2Dial \cite{feng20doc2dial}. 
We test and generate additional dialogs with both the seen documents and unseen documents. 
We augment the original dataset and train it on a downstream model.
The results show that our method improves the performance of the downstream model after augmentation.
We also test scenarios of low-resource settings. We train and evaluate the generative models with only $25\%, 50\%, 75\%$ data. Experimental results show that our method perform well even when training data is scarce.



\begin{figure*}[ht]
    \centering
    \includegraphics[width=\textwidth]{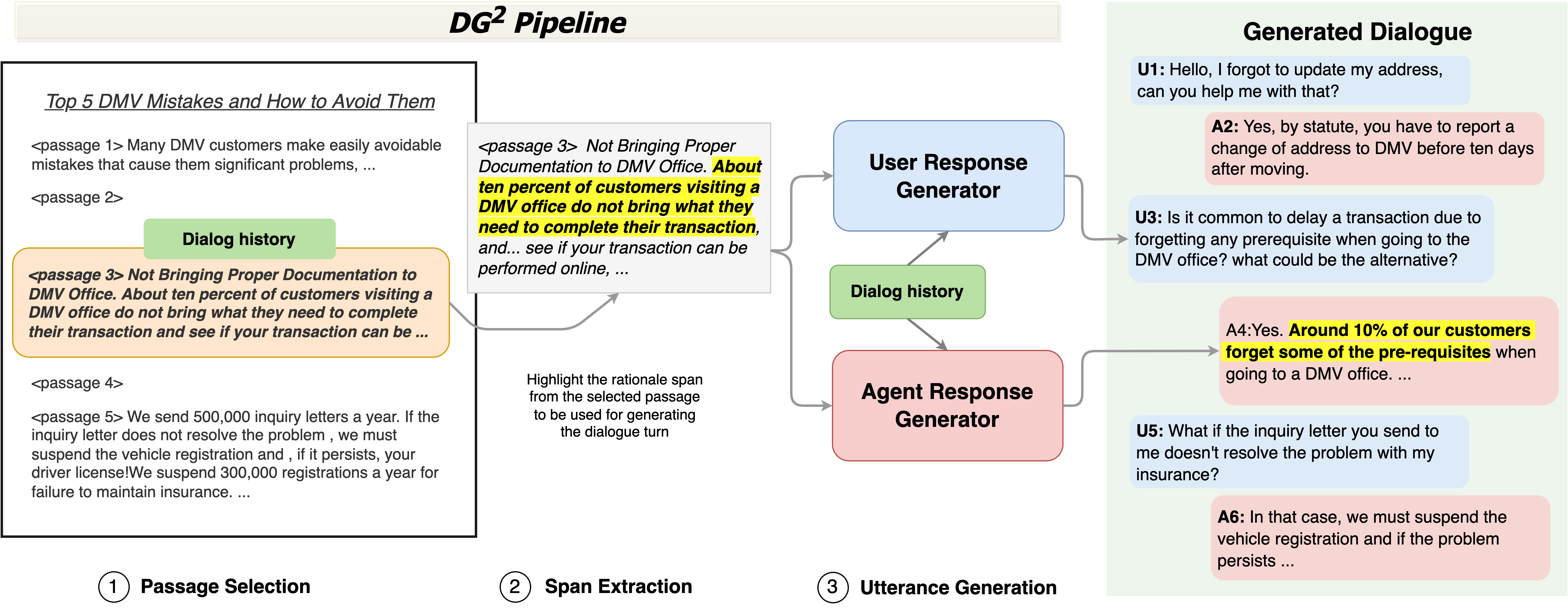}
    \caption{Overall pipeline of $DG^2$. Given a document and the dialogue history, our method iteratively performs  (1) passage selection, (2) rationale extraction, and (3) utterance generation to produce a completed dialogue.}
    \label{fig:pipeline}
\end{figure*}

\section{Related Work}

\subsection{Document Grounded Dialogue Systems}

Document Grounded Dialogue System (DGDS) is the type of dialogue systems that the dialogues are grounded on the given documents.
It helps humans to better retrieve information they want as most of human knowledge is stored in the form of documents.
The study of DGDS can greatly impact the future way of interacting with knowledge.

Recently, there are many document grounded dialogue datasets proposed.
Doc2Dial \cite{feng20dataset} is a representative document grounded dialogue dataset which involved human-to-human conversations and focused on real scenarios under social welfare domains.
Previous datasets such as CoQA \cite{reddy19coqa} and QuAC \cite{choi18quac} focused on machine reading comprehensions.
SharC \cite{saeidi18sharc} is close to Doc2Dial. Its conversations are grounded to short text snippets, and contains follow-up questions.
ABCD \cite{chen21abcd} supports customer service interactions by providing Agent Guidelines as additional documentation to aid in task-oriented conversations.

An example of DGDS from Doc2Dial is shown in Figure 1.
For each turn, the agent needs to look at the specific paragraph inside the document to be capable of answering the user's questions.
Moreover, the agent can also ask follow-up questions.
For A3, the agent asks ``Would you like to know if you are eligible?".
In this way, the agent guides the user to center more on the details in the document.
Due to the complexity of Doc2Dial, simulating such dialogues is highly nontrivial.

\subsection{Data Augmentation}
Data augmentation for question answering and dialogue systems has been well-studied in the past. There are two major directions: paraphrasing existing QA pairs from seed data or generating new QA pairs from scratch.

Paraphrasing is a simple and effective technique to augment natural language datasets. It has been widely used in many NLP tasks including  natural language understanding, question answering, and task-oriented dialog systems \cite{gao20paraphrase} to improve the downstream models' performance. In question answering, paraphrasing with back-translation \cite{chadha19bertqa, bornea21translation} is well-studied for datasets such 
as SQUAD \cite{rajpurkar16squad}.

Another approach is generating new question-answer pairs. Early question-answer generation models used rule-based methods \cite{rajpurkar16squad}. 
More recently, neural network-based question-answer pair generation models have been studied.
PAQ \cite{lewis21paq} generates 65 million question-answer pairs based on Wikipedia and trained a retrieval based that.

However, existing approaches have not explored applications for conversational question answering yet, especially for document grounded dialog systems.
Compared to single-turn question answering datasets like SQUAD \cite{rajpurkar16squad}, it involves additional complexity of modeling dialog flow and interconnection naturalness.
Also, instead of only providing an answer span, datasets like Doc2Dial \cite{feng20dataset} have free-form agent responses.
The agent needs to produce natural utterances conditional to the selected rationale.

Also, existing conversational question generation models \cite{DBLP:conf/eacl/GuMYS21} only focused on the quality of generations but did not address the improvement on downstream models.
We design a specific dialog augmentation approach for document-grounded dialog systems. Our work can synthesize the entire conversation, and can be used to improve down-stream task's performance.

\section{Document-Grounded Dialogue Setup}

A dialogue can be thought of as a series of turns between two interlocutors.  
Within goal-oriented dialogues, we refer to the first speaker as the user, and the second speaker as the agent, whom we model as  $D_{goal} = [(u_1, a_1), (u_2, a_2), ... (u_t, a_t)]$. For the chit-chat setting, the participants are interchangeable so we refer to them as simply User A and User B.  $D_{chat} = [(U_1^a, U_1^b), (U_2^a, U_2^b), ... (U_t^a, U_t^b)]$
In a document-grounded setting, the conversation revolves around the topics and entities mentioned in the associated document.  A document is composed of a series of text passages, which are themselves broken down further into spans.  

Dialogue success is determined by following the typical success metrics for any given task, where the only difference is that the outcome of the conversation is likely to depend on the ability to reason about the contents of the document.  While sophisticated architectures are certainly capable of improving document-grounding, we take a data-centric approach instead by generating new dialogues from the documents to serve as additional training data for the downstream model.

\section{Data Augmentation via $DG^2$}
We propose \textbf{D}ocument-\textbf{G}rounded \textbf{D}ialogue \textbf{G}eneration ($DG^2$) as a method of data augmentation.  
We aim to generate a complete and coherent dialogue given a document by building two bots talking to each other.

Given a document $C$, we can model a dialog $d$ between the user and the agent with:
\begin{equation}
    p(d | C) = \prod_{i=1}^t p(u_i, a_i | c_i \in C) 
\end{equation}
where $u_i$ is the user turn utterance, $a_i$ is the agent turn utterance, and $c_i$ is the selected passage at $i$-th turn.

We further decompose the model into three parts: passage selection, rationale extraction, and utterance generation.
We also apply a filtering model to ensure the quality of generated utterances.


\subsection{Passage Selection} 

A document can often be very long, so it must be divided into smaller passages first. 
Then, we need to rank the passages, and select a relevant passage given the dialogue context. 
We can maximize the passage probability for $c_t$  with contrastive loss where the positive passages are from ground truth, and the negative passages are from the same document.
\begin{equation}
    p(c_t | \{u_i,a_i \}_{i<t},  C)
\end{equation}
During generation, we sample from the probability distribution to select the passage. 
We choose to sample rather than perform greedy selection since this allows for choosing different passages given the same dialogue context, thereby increasing the diversity of the augmentation.

\subsection{Rationale Extraction} 
Next, we further extract a rationale span from the selected passage.
\begin{equation*}
    p(r_t | \{u_i,a_i \}_{i<t}, c_t)
\end{equation*}
Span extraction systems typically model the start and end position of a span independently as $p(r_\text{start} | c) \times (r_\text{end} | c)$.  
This settings works well when the span is short, as is often the case for standard question answering tasks.
However, the spans encountered in some document-grounded dialog datasets are much longer causing problems in traditional approaches.
As an alternative, we propose an autoregressive method that samples the start and end position in sequentially with:
\begin{equation}
p(r_t) = p(r_\text{start} | c) \times  p(r_\text{end} | r_\text{start}, c)
\end{equation}
To ensure that the autoregressive property holds, we add the predicted start position's hidden state $H_\text{start}$ and each position's hidden state $H_i$, and then we project the combined hidden state with a learnable function $f_r$ to get the final predicted end position.
Thus, the training objective \vphantom{for predicting the end position} becomes to maximize 
\begin{equation}
    r_\text{end} = \argmax_{i} f_r (H_\text{start} + H_i)
\end{equation}
When extracting a rationale, we first sample a start position from top-k options.  
Conditioned on this start index, we then sample the end position. 
This allows us to extract different rationales given the same context, which greatly improves the diversity of generated dialogues.

\subsection{Utterance Generation}
Given the selected passage and the extracted rationale, we can now start to generate the user utterance and the agent utterance. 

\paragraph{User Utterance}  As seen in Figure 2, user model generates a user utterance conditioned on the dialog history and the extracted rationale.
Instead of only using the rationale to generate utterances, we provide the context passage along with the rationale for better performance.
To tell the model where the rationale is in the passage, we highlight the rationale span by wrapping its text in the input with ``[" and ``]".
The new passage with the rationale span information is defined as $c'_t$.

We then model the user utterance with a encoder-decoder where the input is the dialogue history and the passage $c'_t$, and the output is the user utterance.
\begin{equation}
    p(u_t) = p(u_t | \{u_i, a_i \}_{i < t}, c'_t)
\end{equation}

\paragraph{Agent Utterance} 
Similar to user utterance generation, we model the agent utterance with a encoder-decoder.
The difference is that the dialogue history now includes the previous generated user utterance.
\begin{equation}
    p(a_t) = p(a_t | \{a_i, u_i \}_{i < t}, c'_t)
\end{equation}

after we have the user utterance, we want to generate a agent response.


\subsection{Filtering the Augmented Data}
Roundtrip consistency checking \cite{alberti19synthetic, zhong20gazp} has previously been used to improve the correctness of generated augmentation data. It utilizes a model to double-check whether the answer span is the same as the span used to generate the question.
Based on this insight, rather than tuning a sampling temperature to trade-off against noise and diversity, we instead greedily pick the rationale span and use consistency checking to filter for quality. For our purposes, we expect the extracted rationale to be aligned with the dialogue context as well as the user utterance.

We build a new passage selector and rationale extraction model such that:
\begin{align}
    &p(\hat{c}_t | \{u_i,a_i \}_{i<t}, , u_t, C) \\
    &p(\hat{r}_t | \{u_i,a_i \}_{i<t}, u_t, \hat{c}_t)
\end{align}
where $\hat{c}_t$ is the predicted passage from the document $C$ with the dialogue context and the generated user utterance, and $\hat{r}_t$ is the prediction rationale within $\hat{c}_t$. When $\hat{r}_t$ is not aligned the previous $r_t$, we remove this utterance $u_t$.  Because rationale spans can be very long, filtering based on exact match will be too strict, so we filter based on f1 word overlap.


\subsection{Document Positional Information} 
When a document is divided into passages, it loses positional information between different passages.
As a dialogue progresses, we can expect to focus more on the later part of a document, which involves more details of a topic.
Therefore, it is important to incorporate the turn information and the passage position information into the model.



We use a simple yet effective method to combine the dialogue turn positional information and passage positional information.
For the speaker positions 
we use a prompt ``user\{num\}:" or ``agent\{num\}:", where ``num" is replaced with the number of turns so far. This allows the model to track how many turns have passed, leading to a more coherent dialog structure.
For the passage positions, we embed a passage index to indicate the location of the passage within the document. Combining the two flows together, the model is able to have conversations focused on the beginning of the document at the first, and naturally shift towards the end of document later.

\begin{table*}[ht]
    \begin{center}
    \resizebox{0.8\textwidth}{!}{
        \begin{tabular}{l|ccc|ccc|c}
            \toprule
             \multirow{2}{*}{Model}  & \multicolumn{3}{c|}{Validation} & \multicolumn{3}{c|}{Test} & \multirow{2}{*}{Span Coverage}  \\
             & EM & F1 & BLEU & EM & F1 & BLEU & \\
             \midrule
             Original data & 58.13 & 72.61 & 37.08 & 58.34 & 73.25 & 36.89 & 48.27  \\
             \, + EDA & \textbf{60.40} & 74.30 & 37.72 & 59.71 & 73.62 & 37.63  & 48.27*  \\ 
             \, + Back-translation & 60.15 & 73.74 & 36.68 & 60.17 & 73.35 & 37.32  & 48.27*  \\
             \, + Paraphrase & 59.97 & 73.92 & 37.76 & 57.98 & 72.71 & 38.40  & 48.27*  \\
             \, + $DG^2$ & 60.30 & \textbf{74.34} & \textbf{38.07} & \textbf{60.92} & \textbf{74.53} & \textbf{38.57} & \textbf{57.65}  \\
             \bottomrule
        \end{tabular}
    }
    \end{center}
    \caption{Experimental results on the Doc2Dial dataset. EM stands for Exact Match. $DG^2$ outperforms all other data augmentation methods on almost every metric.  *EDA, Back-translation, and Paraphrase do not modify span information and thus are unable to increase span coverage in relation to the original data.}
    \label{tab:main_results}
\end{table*}

\section{Experiments}

We first introduce the datasets evaluated with our method, then the baselines for comparisons, and in the end our method's implementation details.

\subsection{Datasets}

\begin{table}[H]
    \begin{center}
    \resizebox{0.48\textwidth}{!}{
    \begin{tabular}{l|cccc|cc}
        \toprule
          & \multicolumn{4}{c|}{Dialogue Level} & \multicolumn{2}{c}{Document Level} \\
          & \#dial & \#turns & \#tok & span & \#doc &  \#tok  \\
         \midrule
         train & 3,474 & 11.8 & 15.0 & 26.5 & 415 & 834  \\
         valid & 661 & 12.1   & 15.3 & 25.8 & 273 & 821  \\
         test  & 661 & 12.0   & 14.9 & 24.5 & 273 & 809  \\
         $DG^2$ & 3,474 & 12.0 & 14.2 & 42.2 & 415 & 834 \\
         \bottomrule
    \end{tabular}
    }
    \end{center}
    \caption{Doc2Dial dataset statistics.  The following abbreviations are made: `dial' is short for dialogue, `tok' is short for tokens, and `doc' is short for documents.}
    \label{tab:doc2dial_statistics}
\end{table}

\paragraph{Doc2Dial} consists of two subtasks around identifying relevant spans based on dialogue context and producing cohesive responses based on extracted rationales \citep{feng20doc2dial}.
Formulated as a span selection task, user utterance understanding requires an agent to interpret user queries in the context of the dialogue history and then select the relevant span from the associated document. Predicted spans are graded based on Exact match (EM) and F1-score.  Exact match is when the predicted span exactly lines up with the actual span.  F1-score balances the recall and precision of the predicted uni-grams compared to the gold span.

The second subtask is agent response prediction, which requires an agent to generate a natural language response to the user query given the dialogue context and the document.   Response quality is measured by SacreBLEU metric \cite{post18sacrebleu} which aims to capture how closely the predicted response lines up with the gold response.  Table~\ref{tab:doc2dial_statistics} shows Doc2Dial's dialogue-level statistics and document-level statistics.



\subsection{Baselines}

We compare against a number of baselines typically used to augment natural language data.  In contrast to our technique, these methods all operate on the existing dialogues, whereas our method generates new dialogues from scratch from the associated document. 

\paragraph{Easy Data Augmentation} \citet{wei19eda} propose to augment data through a series of surface form alterations.  In particular, Easy Data Augmentation (EDA) consists of inserting new tokens, deleting random tokens, swapping pairs of tokens, or replacing tokens with their synonyms.

\paragraph{Back-translation} Back-translation is another strong augmentation method which first translates some text into a separate language and then back-translates to the original language.  We follow BERT-QA \cite{chadha19bertqa}, in translating all user utterances to French and then back to English to augment the original dialogues.

\paragraph{Paraphrase} Paraphrasing can be achieved by training a sequence-to-sequence model on parallel paraphrase pairs corpora.  In particular, we train a BART-base model \cite{lewis20bart} on the MRPC \cite{dolan2005mrpc}, QQP \cite{iyer2017qqp} and PAWS \cite{zhang2019paws} datasets.


\subsection{Coverage Metric}
During inference, any section within the document is fair game for discussion. 
A model trained on dialogues that cover larger portions of the given documents should therefore perform better later on.  
Consequently, a strong data augmentation method should aim to generate dialogues that cover as much of the document as possible.  We formalize this intuition with the span coverage metric, which we calculate as: 

\begin{equation*}
    \text{Coverage} = \frac{\sum_{\text{span}}  |\bigcup_{d \in \text{doc}_i} \bigcup_{s \in d} s|}{|\text{document}_i|}
\end{equation*}
where $s$ refers to spans within a document and $doc$ refers to the number of documents in the corpus.

\subsection{Implementation Details}

For passage ranker, and rationale extraction model, we fine-tuned RoBERTa-base \cite{DBLP:journals/corr/abs-1907-11692} on the downstream training datasets.
For utterance generators, we fine-tuned BART-base \cite{DBLP:conf/acl/LewisLGGMLSZ20}.
We set total input length of 512-tokens  which is 128 tokens for dialogue followed by 360 tokens for the document, with some room left over for special tokens. The augmented data is generated with beam size $4$, top-p $0.9$, and temperature $0.9$.
When utilizing the augmented data, we pre-trained the downstream model on the augmented data for one epoch before fine-tuning \cite{alberti19synthetic}.
The default f1 threshold is set to $0.9$, which we determined by validating against the dev set. For fine-tuning, we train for five epochs, and use the same optimizer of AdamW \cite{loshchilov19adamw} and learning rate of $3e-5$ for all experiments.

\begin{table*}[H]
    \begin{center}
    \resizebox{0.85\textwidth}{!}{
        \begin{tabular}{l|ccc|ccc|ccc}
            \toprule
             Model & \multicolumn{3}{c|}{25\%} & \multicolumn{3}{c|}{50\%} & \multicolumn{3}{c}{75\%} \\
             & EM & F1 & BLEU & EM & F1 & BLEU & EM & F1 & BLEU \\
             \midrule
             Baseline & 43.08 & 64.01 & 32.76 & 41.61 & 62.25 & 34.35 & 58.03 & 72.61 & 36.48 \\
             \, + EDA & 46.68 & 64.68 & 33.97 & \textbf{56.09} & 70.51 & 35.84 & \textbf{59.84} & \textbf{73.40} & 36.24  \\
             \, + Back-translation & \textbf{47.48} & 65.18 & 33.00 & 54.44 & 69.52 & 35.30 & 58.66 & 72.75 & 36.08  \\
             \, + $DG^2$ & 46.48 & \textbf{65.58} & 32.90 & 54.51 & \textbf{71.40} & 35.74 & 58.89 & 73.38 & \textbf{37.01} \\
             \bottomrule
        \end{tabular}
    }
    \end{center}
    \caption{Experimental results on low-resource settings.}
    \label{tab:low_resource}
\end{table*}

\begin{table*}[ht]
    \centering
    \resizebox{0.99\textwidth}{!}{
    \begin{tabular}{c|p{14cm}}
    \hline
    \textbf{Speaker} & \hspace{5cm}\textbf{Utterance} \\
    \hline
    \rowcolor{mygray}
User & I'm thinking about getting some retirement benefits. \\
Agent & Are you wondering what age to start your benefits? \\
&   \textcolor{gray}{\textit{Rationale: What Is The Best Age To Start Your Benefits?}} \\
    \rowcolor{mygray}
User & Yes, I want to know what age I should start my benefits. \\
Agent & Are you planning to apply for Medicare? \\
&   \textcolor{gray}{\textit{Rationale: Should I apply for Medicare?}} \\
    \rowcolor{mygray}
User & Yes, I am. \\
Agent & Then you should know that Medicare usually starts when you reach age 65. \\
&   \textcolor{gray}{\textit{Rationale: Remember, Medicare usually starts when you reach age 65. If you decide to delay starting your benefits, be sure to contact Social Security about 3 months before you turn age 65 to check about applying for Medicare. Even if you have health insurance through a ... }} \\
    \rowcolor{mygray}
User & What if I need to enroll but don't? \\ 
Agent & Your Medicare coverage may be delayed and cost more. \\
&   \textcolor{gray}{\textit{Rationale: your Medicare coverage may be delayed and cost more. Please read the general and special enrollment period information on page 12 of our Medicare booklet to find out what may happen if you delay.}} \\
    \rowcolor{mygray}
User & What age is the best age to begin benefits? \\
Agent & The answer is that there is no one " best age" for everyone and, ultimately, it is your choice. You should make an informed decision about when to apply for benefits based on your individual and family circumstances. \\
&   \textcolor{gray}{\textit{Rationale: The answer is that there is no one " best age " for everyone and, ultimately, it is your choice. You should make an informed decision about when to apply for benefits based on your individual and family circumstances.}} \\
    \hline
    \end{tabular}
    }
    \caption{Generated dialogue showing typical agent and user interaction.  Unlike question answering datasets, the agent can also ask questions to the user to guide the direction of the conversation.}
    \label{tab:doc2dial_good_example}
\end{table*}

\begin{table*}[ht]
    \centering
    \resizebox{0.99\textwidth}{!}{
    \begin{tabular}{c|p{14cm}}
    \hline
    \textbf{Speaker} & \hspace{5cm}\textbf{Utterance} \\
    \hline
    \rowcolor{mygray}
User & I was planning to move and forgot to update my address. What do I do? \\
Agent & By statute, you must report a change of address to DMV within ten days of moving. That is the case for the address associated with your license, as well as all the addresses associated with each registered vehicle, which may differ \\
&   \textcolor{gray}{\textit{Rationale: By statute, you must report a change of address to DMV within ten days of moving. That is the case for the address associated with your license, as well as all the addresses associated with each registered vehicle, which may differ.}} \\
    \rowcolor{mygray}
User & Am I able to do anything if I forget to update my address?
 \\
Agent & It is not sufficient to only: write your new address on the back of your old license; tell the United States Postal Service; or inform the police officer writing you a ticket.
 \\
&   \textcolor{gray}{\textit{Rationale: It is not sufficient to only: write your new address on the back of your old license; tell the United States Postal Service; or inform the police officer writing you a ticket.}} \\
    \rowcolor{mygray}
User & What happens if I forget to update my address?
 \\
Agent & You will miss a suspension order and may be charged with operating an unregistered vehicle and/or aggravated unlicensed operation, both misdemeanors. This really happens, but the good news is this is a problem that is easily avoidable. 
 \\
&   \textcolor{gray}{\textit{Rationale: you will miss a suspension order and may be charged with operating an unregistered vehicle and/or aggravated unlicensed operation, both misdemeanors. This really happens, but the good news is this is a problem that is easily avoidable. Learn more ...}} \\

    \hline
    \end{tabular}
    }
    \caption{A bad generated dialogue here shows repetition. Although the user utterances' surface form are different, their semantic meaning is the same.}
    \label{tab:doc2dial_bad_example}
\end{table*}

\section{Results and Analysis} 

This section shows the results for the full dataset and low-resource settings.
Afterwards, we discuss the results by analyzing generated examples.

\subsection{Main Results}

As shown in Table \ref{tab:main_results}, $DG^2$ achieves the overall best performance compared to other baselines that only augment the original human-annotated data.
Other baselines all show some improvements over the downstream model only trained using the original data.
EDA has very high EM and F1 scores for the rationale extraction task, but suffers at producing coherent dialogues as measured by BLEU.
Paraphrase has relatively lower EM and F1 scores, but it achieves better BLEU scores than EDA and Back-translation.
We suspect that this is because Paraphrase contains more diverse utterances as the inputs than other baselines.

When evaluating the augmented dialogues with the original training set's documents, we find that $DG^2$ achieves higher span coverage.  Unlike the other methods, $DG^2$  is able to generate novel rationales to increase the diversity of the augmented data, which we believe plays a large factor in improving downstream metrics.

\begin{table}[H]
    \begin{center}
    \resizebox{0.36\textwidth}{!}{
    \begin{tabular}{c|c|cc}
        \toprule
         Filtering & \#Spans & EM & F1  \\
         \midrule
         None & - & 57.78 & 73.27  \\
         f1 < 0.5 & top-1 & 57.73 & 73.01  \\
         f1 < 0.9 & top-10 & 58.23 & 73.05 \\
         f1 < 0.9 & top-1 & \textbf{60.80} & \textbf{74.38} \\
         f1 < 0.95 & top-1 & 59.21 & 74.00 \\
         f1 < 0.98 & top-1 & 59.26 & 73.84 \\
         \bottomrule
    \end{tabular}
    }
    \end{center}
    \caption{We test different quality thresholds to determine the optimal level of filtering.  A higher F1-score means that more samples are filtered.}
    \label{tab:filtering}
\end{table}

\subsection{Low Resource Setting}

To further illustrate the performance of $DG^2$, we train all the models with only $25\%, 50\%, 75\%$ of the original training data. We generate the dialogues based on the documents in the knowledge base. In this limited data setting, our model generally outperformed Back-translation. However, compared to EDA, there is still some performance gap.  We suspect that this is because when training with less data, the generative models' performance degenerates faster than the downstream model.
We hope to overcome these issues with further improvements on data quality filtering.


\begin{table*}[ht]
    \begin{center}
    \resizebox{0.85\textwidth}{!}{
        \begin{tabular}{l|ccc|ccc|ccc}
            \toprule
             Model & \multicolumn{3}{c|}{25\%} & \multicolumn{3}{c|}{50\%} & \multicolumn{3}{c}{75\%} \\
             & EM & F1 & BLEU & EM & F1 & BLEU & EM & F1 & BLEU \\
             \midrule
             Baseline & 43.08 & 64.01 & 32.76 & 41.61 & 62.25 & 34.35 & 58.03 & 72.61 & 36.48 \\
             \, + EDA & 46.68 & 64.68 & 33.97 & \textbf{56.09} & 70.51 & 35.84 & \textbf{59.84} & \textbf{73.40} & 36.24  \\
             \, + Back-translation & \textbf{47.48} & 65.18 & 33.00 & 54.44 & 69.52 & 35.30 & 58.66 & 72.75 & 36.08  \\
             \, + $DG^2$ & 46.48 & \textbf{65.58} & 32.90 & 54.51 & \textbf{71.40} & 35.74 & 58.89 & 73.38 & \textbf{37.01} \\
             \bottomrule
        \end{tabular}
    }
    \end{center}
    \caption{Experimental results on low-resource settings.}
\end{table*}

\subsection{Different Filtering Strategies}
Prior works in data augmentation have shown that filtering the synthetically generated examples can provide a meaningful boost in the data quality \citep{chen21gold}.  As a result, we tune against different F1-score thresholds and span counts on the validation set.  When the generated dialogue produces a higher F1-score, then this example is more likely to also produce better results during testing. The span count determines how many examples we consider when calculating this score. While raising the F1-score threshold increases the potential quality of the data, it comes as the expense of keeping fewer of the generated examples. Based on Table \ref{tab:filtering}, we observe a sweet spot at 0.9, where a stricter filtering process would remove too many examples while a looser filtering process would lower the quality too much.


\subsection{Qualitative Analysis}
We now compare and constrast two examples generated by our procedure. Table~\ref{tab:doc2dial_good_example} shows a good example from the document-grounded dialogue dataset.  In the first four turns, the agent guides the user's focus by asking relevant questions.  When the user wants to know more details, the agent then switches to provide the relevant knowledge retrieved from the rationale.  This behavior is different from traditional question answering datasets where the agent simply reacts to user requests rather than exhibiting proactive behavior. On the flipside, one major problem of the current approach is repetition. As shown in Table~\ref{tab:doc2dial_bad_example}, the user continues to ask about forgetting to update their address despite attempts by the agent to answer their query.  Although the surface form of the user utterances are different, the semantic meaning remains the same.  This repetition confuses the agent who then extracts irrelevant rationales, further exacerbating the situation.


\section{Conclusion}
To address the problem of limited data in document-grounded dialogue systems, we propose $DG^2$ to perform data augmentation via dialogue generation. 
Our technique generates diverse utterances grounded on the given document, while filtering the utterances to ensure quality and correctness when training on the downstream model.
We demonstrated the effectiveness of our pipeline by showing the improvement over the previous data augmentation methods.
We additionally show competitive results in the low-resource setting when limited amounts of human annotated data is available for training.  Future work will explore more techniques to filtering for data quality.  We hope this spurs further research into document-grounded augmentation techniques for dialogue systems.

\bibliography{anthology,custom}
\bibliographystyle{acl_natbib}

\appendix



\end{document}